\documentclass[a4paper]{article}

\usepackage{INTERSPEECH_v2}

\usepackage{mathtools}
\usepackage{algorithm}
\usepackage[noend]{algpseudocode}
\usepackage{graphicx}
\usepackage{caption}
\usepackage{subcaption}
\graphicspath{ {imgs/} }

\newcommand{\Dsymbol}[1] {\mathcal{D}_{\mathrm{#1}}}
\newcommand{\KLD}[2] {\Dsymbol{KL}\big( #1 \| #2 \big)}
\newcommand{\JSD}[2] {\Dsymbol{JS}\big( #1 \| #2 \big)}
\newcommand{\E}[2] {\mathbb{E}_{#1}\big[ #2 \big]}
\newcommand {\I} {\mat{I}}      % Identity matrix
\newcommand {\N} {\mathcal{N}}  % Gaussian distribution

\newcommand{\norm}[1] {\| #1 \|}

\newcommand{\pxzy} {\p(\x|\z, \y)}
\newcommand{\ptData} {p^{*}_t}
\newcommand{\ptConv} {p_{t \mid s}}
\newcommand{\psData} {p^{*}_s}

\newcommand{\xn} {\vec{x}_n}
\newcommand{\yn} {\vec{y}_n}
\newcommand{\zn} {\vec{z}_n}
\newcommand{\xhat} {\hat{\vec{x}}}
\newcommand{\xsn} {\x_{s, n}}

\newcommand{\Enc} {\mathcal{E}_{\phi}}
\newcommand{\G} {\mathcal{G}_{\theta}}
\newcommand{\D} {\mathcal{D}_{\psi}}

\newcommand{\q} {q_{\phi}}
\newcommand{\p} {p_{\theta}}

\newcommand{\x} {\vec{x}}
\newcommand{\y} {\vec{y}}
\newcommand{\z} {\vec{z}}

\newcommand {\fpOne} {\mathcal{E}_{\phi_1}}
\newcommand {\fpTwo} {\mathcal{E}_{\phi_2}}

\newcommand{\xObeysPdata} {\x \sim p_t^{*}}
\newcommand{\xObeysPfake} {\x \sim \ptConv}
\newcommand{\zObeysQ} {\z \sim \q}

\newcommand{\SPlen} {log{SP}_{en}}

\title{
  Voice Conversion from Unaligned Corpora using Variational Autoencoding Wasserstein Generative Adversarial Networks}
\name{
  Chin-Cheng Hsu$^1$, 
  Hsin-Te Hwang$^1$, 
  Yi-Chiao Wu$^1$, 
  Yu Tsao$^2$, and
  Hsin-Min Wang$^1$}
\address{
  $^1$Institute of Information Science, Academia Sinica, Taiwan\\
  $^2$Research Center for Information Technology Innovation, Academia Sinica, Taiwan}
\email{
  \{jeremycchsu, hwanght, tedwu, whm\}@iis.sinica.edu.tw,
  yu.tsao@citi.sinica.edu.tw}

\begin{document}
\maketitle

\begin{abstract}
Building a voice conversion (VC) system from non-parallel speech corpora is challenging 
but highly valuable in real application scenarios.
In most situations, the source and the target speakers do not repeat the same texts or they may even speak different languages.
In this case, one possible, although indirect, solution is to build a generative model for speech.
Generative models focus on explaining the observations with latent variables instead of learning a pairwise transformation function, thereby bypassing the requirement of speech frame alignment.
In this paper, we propose a non-parallel VC framework with a variational autoencoding Wasserstein generative adversarial network (VAW-GAN) that explicitly considers a VC objective when building the speech model.
Experimental results corroborate the capability of our framework for building a VC system from unaligned data,
and demonstrate improved conversion quality.

\end{abstract}

% ==============================================================
\section{Introduction}
\label{intro}

The primary goal of voice conversion (VC) is to convert the speech from a source speaker to that of a target, without changing the linguistic or phonetic content.
However, consider the case of converting one's voice into that of another who speaks a different language.
Traditional VC techniques would have trouble dealing with such cases because most of them require \textit{parallel} training data in which many pairs of speakers uttered the same texts.
In this paper, we are devoted to bridging the gap between \textit{parallel} and \textit{non-parallel} VC systems.

\begin{figure}
\centering
  % the percentage right after bracket works like a charm!
  \begin{subfigure}[b]{.25\textwidth}
  \centering
    \includegraphics[width=0.5\linewidth]{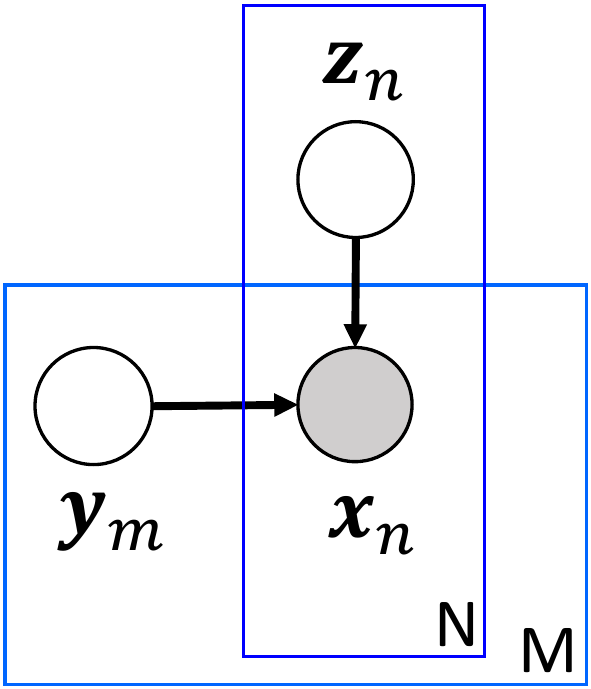}
    \caption{Speech model}
    \label{fig:pgm-plate}
  \end{subfigure}%
  \begin{subfigure}[b]{.25\textwidth}
  \centering
    \includegraphics[width=0.5\linewidth]{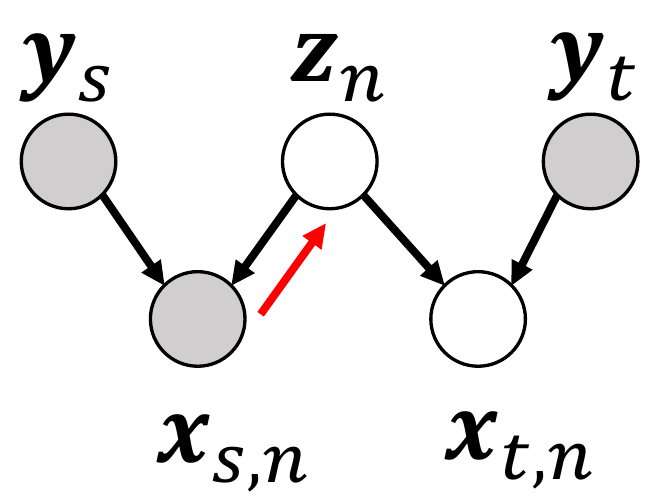}
    \caption{Voice conversion}
    \label{fig:pgm-unroll}
  \end{subfigure}
  \begin{subfigure}[b]{.5\textwidth}
  \centering
    \includegraphics[width=0.6\linewidth]{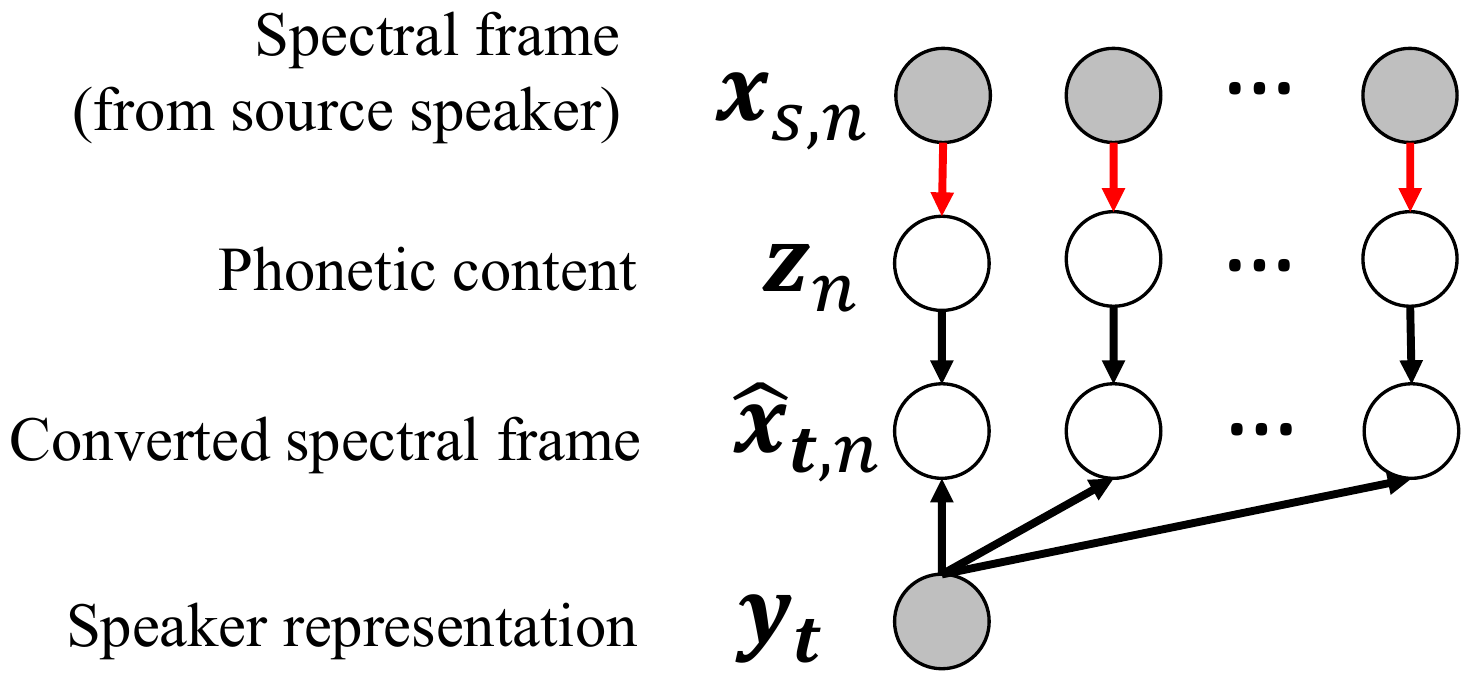}
    \caption{The conversion phase}
    \label{fig:pgm-conv}
  \end{subfigure}
  \caption{
    Graphical models of our voice conversion framework.
    Red arrows represent inference.
    During the conversion phase,
    we first infer the latent variable $\z$ and blend it with the target speaker's representation $\y$.}
\label{fig:pgm}
\end{figure}

We pursue a unified generative model for speech that naturally accommodates VC (Sec.~\ref{sec:model}).
In this framework, we do not have to align any frames or to cluster phones or frames explicitly.
The idea is skeletonized in the probabilistic graphical model (PGM) in Fig.~\ref{fig:pgm}.
In this model, our attention is directed away from seeking alignment.
Rather, what we are concerned about are 1) finding a good inference model (Sec.~\ref{sec:cvae}) for the latent variable $\z$ and 2) building a good synthesizer whose outputs match the distribution of the real speech of the target (Sec.~\ref{sec:gan} through \ref{sec:wgan}).
In this paper, we present a specific implementation in which 
a variational autoencoder (VAE \cite{DBLP:journals/corr/KingmaW13}) assumes the inference task and 
a Wasserstein generative adversarial network (W-GAN \cite{DBLP:journals/corr/ArjovskyCB17}) undertakes speech synthesis.
Our contribution is two-fold:
  \begin{itemize}
    \item We introduce the W-GAN to non-parallel voice conversion and elucidate the reason why it fits this task (Sec.~\ref{sec:model}).
    \item We demonstrate the ability of W-GAN to synthesize more realistic spectra (Sec.~\ref{sec:exp}).
  \end{itemize}

\section{Non-parallel voice conversion via deep generative models}
\label{sec:model}

Given spectral frames $X_s=\{\x_{s, n}\}^{N_s}_{n=1}$ from the source speaker 
and those $X_t=\{\x_{t, n'}\}^{N_t}_{n'=1}$ from the target,
assume that the real data distributions of the source and the target respectively admit a density $\psData$ and $\ptData$.
Let $f$ be a voice conversion function that induces a conditional distribution $\ptConv$.  
The goal of VC is to estimate $f$ so that $\ptConv$ best approximates the real data distribution $\ptData$:
\begin{equation}
  \ptConv(f(\x_{s, n})) \approx \ptData(\x_{t, n'}).
  \label{eq:vc-pdf}
\end{equation}

We can decompose the VC function $f$ into two stages according to the PGM in Fig.~\ref{fig:pgm-unroll}.
In the first stage, a speaker-independent encoder $\Enc$ infers a latent content $\zn$.
In the second stage, a speaker dependent decoder $\G$ mixes $\zn$ with a speaker-specific variable $\y$ to reconstruct the input.
The problem of VC is then reformulated as:
\begin{equation}
  \x_{t, n'} \approx f(\xsn) = \G(\Enc(\xsn), \y).
  \label{eq:vc-ae}
\end{equation}

In short, this model explains the observation $\x$ using two latent variables $\y$ and $\z$.
We will drop the frame indices whenever readability is unharmed.
We refer to $\y$ as the \textit{speaker representation} vector because it is determined solely by the speaker identity.
We refer to $\z$ as the \textit{phonetic content} vector because with a fixed $\y$, we can generate that speaker's voice by varying $\z$.
Note that the term \textit{phonetic content} is only valid in the context of our experimental settings where the speech is natural, noise-free, and non-emotional.

% Point: why we can do NP VC using this model?
This encoder-decoder architecture facilitates VC from unaligned or non-parallel corpora.
The function of the encoder is similar to a phone recognizer whereas the decoder operates as a synthesizer.
The architecture enables voice conversion for the following reasons.
The speaker representation $\y$ can be obtained from training.
The encoder can infer the phonetic content $\z$.
The synthesizer can reconstruct any spectral frame $\x$ with the corresponding $\y$ and $\z$.
Combining these elements, we can build a non-parallel VC system via optimizing the encoder, the decoder (synthesizer), and the speaker representation.
With the encoder, frame-wise alignment is no longer needed;
frames that belong to the same phoneme class now hinge on a similar $\z$.
With this conditional synthesizer, VC becomes as simple as replacing the speaker representation $\y$.
(as illustrated in Fig.~\ref{fig:pgm-conv}).

We delineate our proposed method incrementally in three subsections:
a conditional variational autoencoder (C-VAE) in Sec.~\ref{sec:cvae}, 
a generative adversarial nets (GAN) applied to improve the over-simplified C-VAE model in Sec.~\ref{sec:gan}, and
a Wasserstein GAN (W-GAN) that explicitly considers VC in the training objectives in Sec.~\ref{sec:wgan}.

  \subsection{Modeling speech with a C-VAE}
  \label{sec:cvae}
  Recent works have proven the viability of speech modeling with VAEs
  \cite{DBLP:conf/apsipa/HsuHWTW16,DBLP:conf/interspeech/BlaauwB16}.
  A C-VAE that realizes the PGM in Fig.~\ref{fig:pgm-plate} maximizes a variational lower bound of the log-likelihood:
  \begin{equation}
    \log{\p(\x | \y)} \leq -J_{vae}(\x | \y) = -(J_{obs}(\x | \y) + J_{lat}(x)),
    \label{eq:vlb}
  \end{equation}
  \begin{equation}
    J_{lat}(\phi; \x) =  \KLD{\q(\z|\x)}{\p(\z)},
    \label{eq:vae-lat}
  \end{equation}
  \begin{equation}
    J_{obs}(\phi, \theta; \x, \y) = -\E{\q(\z|\x)}{\log{\p(\x|\z, \y)}},
    \label{eq:vae-obs}
  \end{equation}
  where
  $\x \in X_s \cup X_t$,
  $\Dsymbol{KL}$ is the Kullback-Leibler divergence, 
  $\p(\z)$ is our prior distribution model of $\z$,
  $\p(\x|\z, \y)$ is our synthesis model, and
  $\q(\z|\x)$ is our inference model.
  Note that the synthesis is conditioned on an extra input $\y$, thus the name conditional VAE.

  In order to train the C-VAE, we have to simplify the model in several aspects.
  First, we choose $\pxzy$ to be a normal distribution whose covariance is an identity matrix.
  Second, we choose $\p(\z)$ to be a standard normal distribution.
  Third, the expectation over $\z$ is approximated by sampling methods.
  With these simplifications, we can avoid intractability and focus on modeling the statistics of the Gaussian distributions.

  For the phonetic content $\z$, we have:
  \begin{equation}
  \begin{split}
  \label{eq:sampling}
    \q(\z|\x) & = \N(\z \mid \fpOne(\x),~diag(\fpTwo(\x))),
  \end{split}
  \end{equation}
  where ${\phi_1} \cup {\phi_2} = \phi$ are the parameters of the encoder, and $\fpOne$ and $\fpTwo$ are the inference models of mean and variance.
  For the reconstructed or converted spectral frames, we have:
  \begin{equation}
  \label{eq:vis-gauss}
  \begin{split}
    \p(\x_{m,n}|\z_n, \y_m)  = \N(\x_{m,n} \mid \G(\z_n, \y_m), \I).
  \end{split}   
  \end{equation}

  Training this C-VAE means maximizing (\ref{eq:vlb}).
  For every input $(\xn, \yn)$,
  we can sample the latent variable $\zn$ using the re-parameterization trick described in \cite{DBLP:journals/corr/KingmaW13}.
  With $\zn$ and $\yn$, the model can reconstruct the input,
  and by replacing $\yn$, it can convert voice.
  This means that we are building virtually multiple models in one.
  Conceptually, the speaker switch lies in the speaker representation $\y$ because the synthesis is conditioned on $\y$.

  \subsection{Improving speech models with GANs}
  \label{sec:gan}
  Despite the effectiveness of C-VAE,
  the simplification induces inaccuracy in the synthesis model.
  This defect originates from the fallible assumption that  the observed data is normally distributed and uncorrelated across dimensions.
  This assumption gave us a defective learning objective, leading to muffled converted voices.
  Therefore, we are motivated to resort to models that side-step this defect.

  We can improve the C-VAE by incorporating a GAN objective \cite{DBLP:conf/icml/LarsenSLW16} 
  into the decoder.
  A vanilla GAN \cite{DBLP:journals/corr/GoodfellowPMXWOCB14} consists of two components: 
  a generator (synthesizer) $\G$ that produces realistic spectrum and 
  a discriminator $\D$ that judges whether an input is a true spectrum or a generated one.
  These two components seek an equilibrium in a min-max game with the Jensen-Shannon divergence $\Dsymbol{JS}$ as the objective, which is defined as follows:
  \begin{equation}
  \begin{split}
    ~~& J_{gan}(\theta, \psi; \x) = 2~\JSD{\ptData}{\ptConv} + 2 \log{2}. \\
    =~& \E{\xObeysPdata}{\log{\frac{\ptData}{\ptData + \ptConv}}} 
      + \E{\xObeysPfake}{\log{\frac{\ptConv}{\ptData + \ptConv}}} \\
    =~& \E{\xObeysPdata}{\log{\D^*(\x)}} 
      + \E{\zObeysQ}{\log{\big(1 - \D^*(\G(\z)) \big)}},
     \label{eq:gan}
  \end{split}
  \end{equation}
  where $\D^*$ denotes the optimal discriminator, which is the density ratio in the second equality in (\ref{eq:gan}).
  We can view this as a density ratio estimation problem without explicit specification of distributions.

  Presumably, GANs produce sharper spectra because they optimize a loss function between two distributions in a more direct fashion.
  We can combine the objectives of VAE and GAN by assigning VAE's decoder as GAN's generator
  to form a VAE-GAN \cite{DBLP:conf/icml/LarsenSLW16}.
  However, the VAE-GAN does not consider VC explicitly.
  Therefore, we propose our final model: variational autoencoding Wasserstein GAN (VAW-GAN).

  \subsection{Direct consideration of voice conversion with W-GAN}
  \label{sec:wgan}
  A deficiency in the VAE-GAN formulations is that it treats VC indirectly.
  We simply assume that when the model is well-trained, it naturally equips itself with the ability to convert voices.
  In contrast, we can directly optimize a non-parallel VC loss by renovating $\Dsymbol{JS}$ with a Wasserstein objective \cite{DBLP:journals/corr/ArjovskyCB17}.

    \subsubsection{Wasserstein distance}
    The Wasserstein-1 distance is defined as follows:
    \begin{equation}
      \label{eq:wdist}
      W(\ptData, \ptConv) = 
        \inf_{\gamma \in \Pi(\ptData, \ptConv)} 
          \E{(\x, \xhat) \sim \gamma}{ \norm{\x - \xhat} }
    \end{equation}
    where $\Pi(p, q)$ denotes the set of all joint distributions $\gamma(\x, \xhat)$ whose marginals are respectively $\ptData$ and $\ptConv$.
    According to the definition,
    the Wasserstein distance is calculated from the optimal transport,
    or the best frame alignment.
    Note that (\ref{eq:wdist}) is thus suitable for parallel VC.
    %Note that the optimal transport may be a reasonable alignment conceptually,
    %but it is not necessarily the best alignment for voice conversion.

    On the other hand, the Kantorovich-Rubinstein duality \cite{optimal-transport} of (\ref{eq:wdist}) allows us to explicitly approach non-parallel VC:
    \begin{equation}
      W(\ptData, \ptConv) = \sup_{\|\mathcal{D}\|_L \leq 1} 
      \E{\x \sim \ptData}{\mathcal{D}(\x)} - \E{\x \sim \ptConv}{\mathcal{D}(\x)},
    \end{equation}
    where the supremum is over all 1-Lipschitz functions $\mathcal{D}: \mathcal{X} \to R$.
    If we have a parameterized family of functions $\mathcal{D}_{\psi \in \Psi}$ that are all K-Lipschitz for some K, we could consider solving the problem:
    \begin{equation}
      \max_{\psi \in \Psi} ~ J_{wgan},
    \end{equation}
    where $J_{wgan}$ is defined as
    \begin{equation}
      \E{\x \sim \ptData}{\D(\x)} - \E{\z \sim \q(\z|\x)}{\D(\G(\z), \y_t)}
    \label{eq:wgan}
    \end{equation}
    Alignment is not required in this formulation because of the respective expectations.
    What we need now is a batch of real frames from the target speaker, 
    another batch of synthetic frames converted from the source into the target speaker,
    and a good discriminator $\D$.

    \subsubsection{VAW-GAN}
    Incorporating the W-GAN loss (\ref{eq:wgan}) with (\ref{eq:vlb}) yields our final objective:
    \begin{equation}
    \begin{split}
      J_{vawgan} = & - \KLD{\q(\zn|\xn)}{p(\zn)} \\
                     & + \E{\z \sim \q(\z|\x)}{\log{\p(\x|\z, \y)}} \\
                     & + \alpha~\E{\x \sim \ptData}{\D(\x)} \\
                     & - \alpha~\E{\z \sim \q(\z|\x)}{\D(\G(\z, \y_t))}
    \label{eq:vaw-gan}
    \end{split}
    \end{equation}
    where $\alpha$ is a coefficient which emphasizes the W-GAN loss.
    This objective is shared across all three components: the encoder, the synthesizer, and the discriminator.
    The synthesizer minimizes this loss whereas the discriminator maximizes it; consequently the two components have to be optimized in alternating order.
    For clarity, we summarize the training procedures in Alg. \ref{algo1}.
    Note that we actually use an update schedule for $\D$ instead of training it to real optimality.

    \begin{algorithm}[t]
    \caption{VAE-WGAN training}
      \begin{algorithmic}[0]
      \Function{autoencode}{$X, y$}
        \State $Z_\mu    \leftarrow \fpOne(X)$
        \State $Z_\sigma \leftarrow \fpTwo(X)$
        \State $Z \leftarrow$ sample from $\N(Z_\mu,~Z_\sigma)$
        \State $X^{\prime} \leftarrow \G(Z,~y)$
        \State \Return $X^{\prime},~Z$
      \EndFunction  
      \State
      \State \vec{\phi}, \vec{\theta}, \vec{\psi} $\leftarrow$ initialization
      \While {not converged}
        \State $X_s \leftarrow$ mini-batch of random samples from source
        \State $X_t \leftarrow$ mini-batch of random samples from target
        \State $X^{'}_s, Z_s \leftarrow$ AUTOENCODE($X_s,~y_s$)
        \State $X^{'}_t, Z_t \leftarrow$ AUTOENCODE($X_t,~y_t$)
        \State $X_{t \mid s} \leftarrow \G(Z_s,~y_t)$
        \State $J_{obs}  \leftarrow J_{obs}(X_s) + J_{obs}(X_t)$
        \State $J_{lat}  \leftarrow J_{lat}(Z_s) + J_{lat}(Z_t)$
        \State $J_{wgan} \leftarrow J_{wgan}(X_t,~X_s)$
        \State
        \State // Update the encoder, generator, and discriminator
        \While {not converged}
          \State $\vec{\psi} ~ \xleftarrow{update} ~ -\nabla_\psi (-J_{wgan})$
        \EndWhile

        \State $\vec{\phi}   ~ \xleftarrow{update} ~ -\nabla_\phi (J_{obs} + J_{lat})$
        \State $\vec{\theta} ~ \xleftarrow{update} ~ -\nabla_\theta (J_{obs} + \alpha J_{wgan})$
        
      \EndWhile
      \end{algorithmic}
    \label{algo1}
    \end{algorithm}

\section{Experiments}
\label{sec:exp}

    \subsection{The dataset}
      The proposed VC system was evaluated on the Voice Conversion Challenge 2016 dataset \cite{DBLP:conf/interspeech/TodaCSVWWY16}. 
      The dataset was a parallel speech corpus;
      however, frame alignment was not performed in the following experiments.

      We conducted experiments on a subset of 3 speakers. 
      In the inter-gender experiment, we chose SF1 as the source and TM3 as the target.
      In the intra-gender experiment, we chose TF2 as the target.
      We used the first 150 utterances (around 10 minutes) per speaker for training,
      the succeeding 12 for validation,
      and 25 (out of 54) utterances in the official testing set for subjective evaluations.

    \subsection{The feature set}
      We used the STRAIGHT toolkit \cite{Kawahara99-STRAIGHT} to extract speech parameters, including the STRAIGHT spectra (SP for short), aperiodicity (AP), and pitch contours (F0). 
      The rest of the experimental settings were the same as in \cite{DBLP:conf/apsipa/HsuHWTW16},
      except that we rescaled log energy-normalized SP (denoted by $\SPlen$) to the range of $[-1, 1]$ dimension-wise.
      Note that our system performed frame-by-frame conversion without post-filtering and that we utilized neither contextual nor dynamic features in our experiments.

    \subsection{Configurations and hyper-parameters}
      The baseline system was the C-VAE system (denoted simply as VAE) \cite{DBLP:conf/apsipa/HsuHWTW16}
      because its performance had been proven to be on par with another simple parallel baseline.
      In our proposed system, the encoder, the synthesizer, and the discriminator were convolutional neural networks.
      The phonetic space was 64-dimensional and assumed to have a standard normal distribution.
      The speaker representation were one-hot coded, and their embeddings were optimized as part of the generator parameters\footnote{
        Due to space limitations,
        the rest of the specification of hyper-parameters and audio samples can be found on-line:
        https://github.com/JeremyCCHsu/vc-vawgan}.

    \subsection{The training and conversion procedures}
      We first set $\alpha$ to 0 to exclude W-GAN,
      and trained the VAE till convergence to get the baseline model.
      Then, we proceeded on training the whole VAW-GAN via setting $\alpha$ to 50.

      Conversion was conducted on a frame-by-frame basis as shown in Fig.~\ref{fig:pgm-conv}.
      First, $\Enc$ inferred the phonetic content $\zn$ from $\x_{s, n}$
      Then, we specified a speaker identity (integer, the subscript $t$ in $\y_t$) that retrieved the speaker representation vector $\y$.
      The synthesizer $\G$ then generated a conditional output frame $\xhat$ using $\zn$ and $\y_t$.

  \subsection{Subjective evaluations}
  Five-point mean opinion score (MOS) tests were conducted in a pairwise manner.
  Each of the 10 listeners graded the pairs of outputs from the VAW-GAN and the VAE.
  Inter-gender and intra-gender VC were evaluated respectively.

\begin{figure}
  \centering
     \includegraphics[width=1\linewidth]{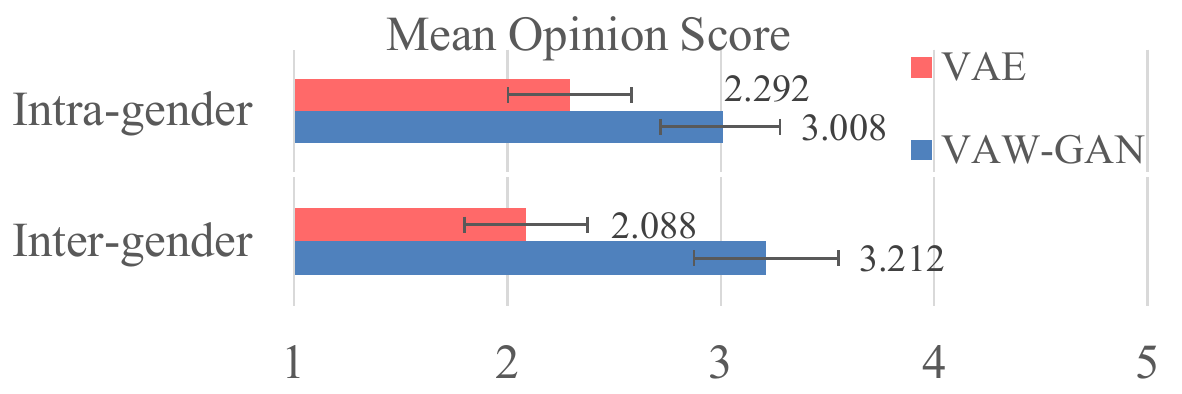}
    \caption{MOS on naturalness. The source is SF1, and the targets are TF2 and TM3.}
    \label{fig:mos}
  \end{figure}

\begin{figure}[t]
    \includegraphics[width=0.42\textwidth]{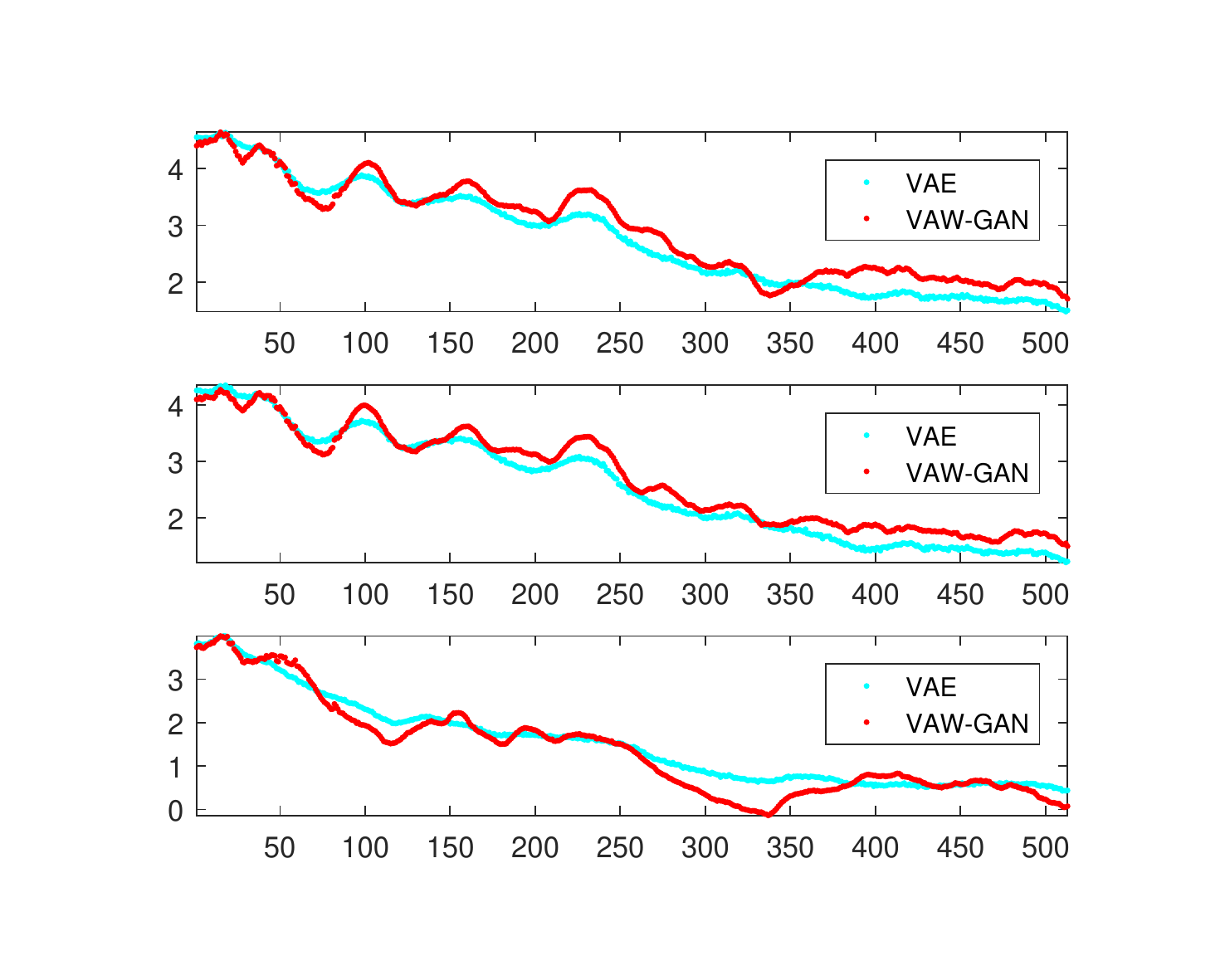}
    \centering
    \caption{Selected frames of the STRAGIHT spectra converted from SF1 to TM3. The spectral envelopes from the VAW-GAN outputs are less smooth across the frequency axis.}
  \label{fig:sp}
  \end{figure}

    The MOS results on naturalness shown in Fig.~\ref{fig:mos} demonstrate that
    VAW-GAN significantly outperforms the VAE baseline (p-value $\ll$ 0.01 in paired t-tests).
    The results are in accordance with the converted spectra shown in Fig.~\ref{fig:sp},
    where the output spectra from VAW-GAN express richer variability across the frequency axis, hence reflecting clearer voices and enhanced intelligibility.

	We did not report objective evaluations such as mean mel-cepstral coefficients because we found  inconsistent results with the subjective evaluations.
    However, similar inconsistency is common in the VC literature because it is highly likely that those evaluations are inconsistent with human auditory systems \cite{Toda07, Chen14-NN-VC, Hwang15-NN-VC}.
    The performance of speaker similarity was also unreported because we found that it remained about the same as that of \cite{Wu16-LLE-VC} (System B in \cite{DBLP:conf/interspeech/WesterWY16}).

  \begin{figure}[t]
    \includegraphics[width=0.42\textwidth]{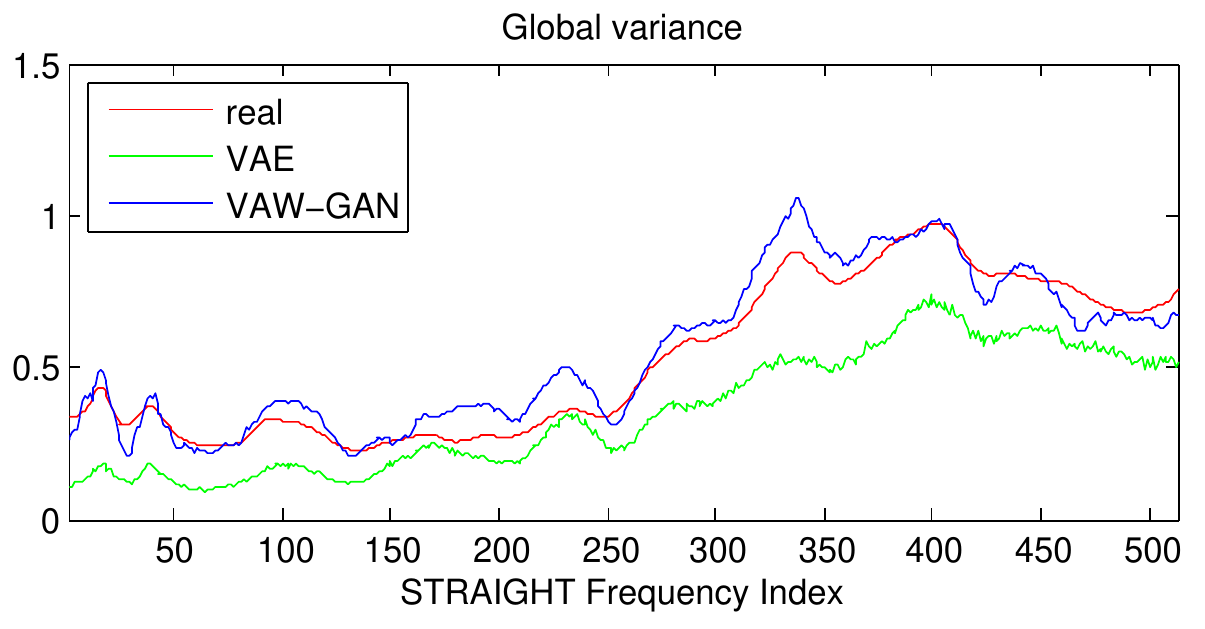}
    \centering
    \caption{The GV computed from the $\SPlen$ over all non-silent frames from speaker TM3.}
    \label{fig:gv}
 \end{figure}

\section{Discussions}
	\subsection{W-GAN improved spectrum modeling}
      As we can see in Fig.~\ref{fig:sp},
      the spectral envelopes of the synthetic speech from VAW-GAN are more structured,
      with more observable peaks and troughs.
      Spectral structures are key to the speech intelligibility,
      indirectly contributing to the elevated MOS.
      In addition, the more detailed spectral shapes in the high-frequency region reflect clearer (non-muffled) voice of the synthetic speech.

     \subsection{W-GAN as a variance modeling alternative}
     The Wasserstein objective in (\ref{eq:vaw-gan}) is minimized when the distribution of the converted spectrum $\ptConv$ is closest to the true data distribution $\ptData$.
     Unlike VAE that assumes a Gaussian distribution on the observation, 
     W-GAN models the observation implicitly through a series of stochastic procedures,
     without prescribing any density forms.
     In Fig.~\ref{fig:gv}, we can observe that the output spectra of the VAW-GAN system have larger variance compared to those of the VAE system.
     The global variance (GV) of the VAW-GAN output may not be as good as that of the data but the higher values indicate that VAW-GAN does not centralize predicted values at the mean too severely.
     Since speech has a highly diverse distribution,
     it requires more sophisticated analysis on this phenomenon.

     \subsection{Imperfect speaker modeling in VAW-GAN}
     The reason that the speaker similarity of the converted voice is not improved reminds us of the fact that both VAE and VAW-GAN optimize the same PGM, thus the same speaker model.
     Therefore, modeling speaker with one global variable might be insufficient.
     As modeling speaker with a frame-wise variable may conflict with the phonetic vector $\z$,
     we may have to resort to other PGMs.
     We will investigate this problem in the future.

\section{Related work}
To handle \textit{non-parallel} VC, many researchers resort to frame-based, segment-based, or cluster-based alignment schemes.
One of the most intuitive ways is to apply an automatic speech recognition (ASR) module to the utterances, and proceed with explicit alignment or model adaptation 
\cite{DBLP:conf/apsipa/DongYLEHMTLL15, DBLP:conf/icassp/ZhangTTW08}.
The ASR module provides every frame with a phonetic label (usually the phonemic states).
It is particularly suitable for text-to-speech (TTS) systems because they can readily utilize these labeled frames
\cite{DBLP:conf/icassp/SongZZ13}.
A shortcoming with these approaches is that they require an extra mapping to realize cross-lingual VC.
To this end, the INCA-based algorithms 
\cite{DBLP:journals/taslp/ErroMB10a, DBLP:conf/icassp/Agiomyrgiannakis16}
were proposed to iteratively seek frame-wise correspondence using converted surrogate frames.
Another attempt is to separately build frame clusters for the
source and the target, and then set up a mapping between them
\cite{DBLP:conf/interspeech/NeySBH04}.

Recent advances include \cite{DBLP:conf/apsipa/WuWX16}, in which the authors exploited i-vectors to represent speakers.
Their work differed from ours in that they adopted explicit alignment during training.
In \cite{DBLP:conf/interspeech/XieSL16}, the authors represented the phonetic space with senone probabilities outputted from an ASR module, and then generated voice by means of a TTS module.
Despite differences in realization, our models do share some similarity ideally.

\section{Conclusions}
We have presented a voice conversion framework that is able to directly incorporate a non-parallel VC criterion into the objective function.
The proposed VAW-GAN framework improves the outputs with more realistic spectral shapes.
Experimental results demonstrate significantly improved performance over the baseline system.

\section{Acknowledgements}
This work was supported in part by the Ministry of Science and Technology of Taiwan under Grant: MOST 105-2221-E-001-012-MY3.

\bibliographystyle{IEEEtran}
\bibliography{jrm1}

\end{document}